\documentclass[runningheads]{llncs}

\usepackage{graphicx}
\usepackage{booktabs}
\usepackage{cite}
\usepackage{amsmath}

\usepackage{xcolor}

\begin{document}
\title{General Purpose Image Encoder DINOv2 for Medical Image Registration}
%
%
\author{Xinrui Song\inst{1} \and Xuanang Xu\inst{1} \and Pingkun Yan*\inst{1}}
\institute{Department of Biomedical Engineering and the Center for Biotechnology and Interdisciplinary Studies\\
Rensselaer Polytechnic Institute, Troy, NY 12180, USA\\
*\email{yanp2@rpi.edu}}

\authorrunning{Song et al.}

\maketitle        
\begin{abstract}

Existing medical image registration algorithms rely on either dataset specific training or local texture-based features to align images. The former cannot be reliably implemented without large modality-specific training datasets, while the latter lacks global semantics thus could be easily trapped at local minima. In this paper, we present a training-free deformable image registration method, DINO-Reg, leveraging a general purpose image encoder DINOv2 for image feature extraction. The DINOv2 encoder was trained using the ImageNet data containing natural images. We used the pretrained DINOv2 without any finetuning. Our method feeds the DINOv2 encoded features into a discrete optimizer to find the optimal deformable registration field. We conducted a series of experiments to understand the behaviour and role of such a general purpose image encoder in the application of image registration.
Combined with handcrafted features, our method won the first place in the recent OncoReg Challenge. To our knowledge, this is the first application of general vision foundation models in medical image registration. 

\end{abstract}
\section{Introduction}

Deformable image registration non-linearly aligns a moving image to a reference image \cite{haskins2020deep}. The task is relevant in treatment planning, atlas-based segmentation, and multimodal image fusion.
Traditionally, deformable registration methods involve an iterative optimization process, in which a quantified similarity metric between the moving and reference images are maximized as the objective. Before the deep learning era, the similarity metrics were based on handcrafted features that can bridge the modality gaps and sample differences between the moving and reference image~\cite{heinrich2012mind}. %
Since the creation of Voxelmorph~\cite{balakrishnan2019voxelmorph}, the latest benchmarking for deformable image registration, many deep learning-based methods~\cite{mok2020large} took a different path of directly predicting a displacement field given the self-learned hierarchical features extracted from the input moving and reference images. However, these two types of methods each suffer from their own limitation. Handcrafted features are often intensity and gradient-based. Although these features are sensitive to corner points and contours, they lack global semantics and therefore may be affected by local minima. Deep learning methods that directly predict the displacement fields, on the other hand, lack explanability and require manual segmentation at training time to overcome multi-modal differences~\cite{balakrishnan2019voxelmorph,mok2020large,hering2022learn2reg}. 

To address the aforementioned drawbacks, a new group of methods use deep neural networks for feature extraction, combined with a subsequent optimizer that establishes the correspondence between the extracted features~\cite{haskins2019learning,li2023samconvex,song2021cross,song2022cross,siebert2021fast}. Deep learning-based features contain rich semantics and are therefore innately explainable. However, existing deep learning-based feature encoders for medical image registration all require modality-specific training. SAMConvex~\cite{li2023samconvex}, for example, is trained on and applied to CT data exclusively. Moreover, deep learning-based encoders require huge datasets to be properly trained. Given the scarcity of medical image data, such methods are not practical in most clinical settings.

Building upon the latest advancements in deep learning, the emergence of foundation models in computer vision has introduced a paradigm shift in approaching complex visual tasks. Self-supervised vision foundation models like DINOv2~\cite{oquab2023dinov2} exemplify this evolution, offering pre-trained models that have learned rich representations from vast unlabeled datasets, thereby mitigating the need for task-specific data in initial training phases. Though pre-trained on natural image datasets, these models excel in understanding global semantics and local details across diverse visual domains, making them highly adaptable for medical image analysis tasks, such as classification~\cite{baharoon2023dinoradiology} and segmentation~\cite{ye2022desd}, beyond their initial training scope. In this work, we will demonstrate that this adaptability can also benefit the tasks of deformable image registration, where capturing the nuanced differences and similarities between images is crucial. 

We propose DINO-Reg, a novel image registration pipeline that utilizes DINOv2 to encode medical image features with rich semantics while requiring no training on different modalities and datasets. The proposed method was used in the winning solution of the recent OncoReg challenge. \footnote{https://learn2reg.grand-challenge.org/} Our main contribution in this paper can be summarized as: (1) the first exploration of leveraging the self-supervised learning model (\emph{i.e.}, DINOv2) fully trained by natural images to extract medical image features for deformable registration; (2) an overall framework design that exploits the full capacity of DINOv2 features; (3) extensive experiments to validate the proposed framework on a wide range of real-world medical image datasets; and (4) proposing a training-free framework for medical image registration.

\section{Method}
\label{sec:method}

\begin{figure}[t]
\includegraphics[width=\textwidth]{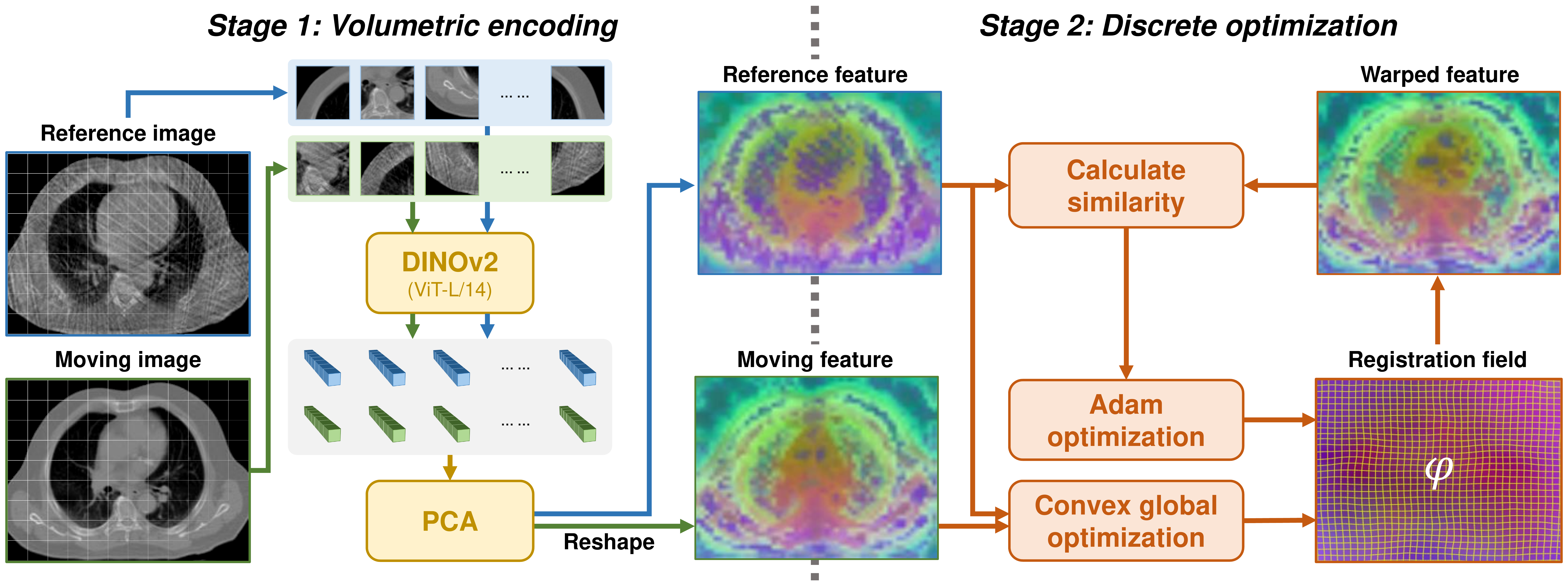}
\caption{Overview of the proposed DINO-Reg framework.} \label{fig1_overview}
\end{figure}

The overall framework of the proposed DINO-Reg is depicted in Fig.~\ref{fig1_overview}. Our algorithm consists of two main parts: volumetric feature encoding with DINOv2~\cite{oquab2023dinov2}, and the feature-based optimization with ConvexAdam~\cite{siebert2021fast}. In the following sections, we will first introduce the preliminary of DINOv2 model (Sec.~\ref{sec:dino_preliminary}) and how we use this 2D model for 3D volumetric feature encoding (Sec.~\ref{sec:vol_feature_extract}), then explain the convex optimization process (Sec.~\ref{sec:ncc_optimization}).

\subsection{Preliminary of DINOv2}
\label{sec:dino_preliminary}

DINOv2 (interpreted as self-\textbf{DI}stillation with \textbf{NO} labels, version 2)~\cite{caron2021dinov1,oquab2023dinov2} is a state-of-the-art self-supervised learning framework that forms the cornerstone of our approach. It leverages the principle of self-distillation to learn knowledge from unlabeled images via a self-supervised manner. Specifically, DINOv2 has a dual-network architecture comprising a teacher and a student network. The two networks share identical network architectures (typically built on Vision Transformers (ViTs)~\cite{dosovitskiy2020vit}) but trained differently. Given an input image, the student and teacher networks are fed with different augmentations of the image, while constrained to extract consistent features. During training stage, the parameters in the student network are optimized through gradient decent algorithms (such as stochastic gradient descent (SGD) and Adam~\cite{kingma2014adam}) while the teacher network's parameters are updated by the moving average of their counterparts in the student network. The trained teacher network is often used as the final product. Such a general purpose foundation model can serve as a powerful feature extractor for the downstream tasks such as classification and segmentation.

In our study, we harness the teacher network of a pretrained DINOv2 model to extract robust and discriminative features from both moving and reference images without any finetuning. The DINOv2 model, pretrained on the ImageNet~\cite{deng2009imagenet} dataset comprising 1.3M unlabeled 2D natural images, is adept at processing 2D slices from volumetric computed tomography (CT) or magnetic resonance (MR) images. The teacher and student networks utilize a ViT-L/14 architecture with a hidden dimension of 1,024. For a monochrome CT slice from the OncoReg dataset, it is initially duplicated to a 3-channel RGB image and resized from 256$\times$192 to 728$\times$560($\times$3 channels). This resized image is divided into 52$\times$40=2,080 non-overlapped patches, each measuring 14$\times$14($\times$3 channels) pixels, subsequently encoded into 2,080 patch tokens with an embedding dimension of 1,024. Prior to input into the ViT-L/14 networks, a class token and four register tokens~\cite{darcet2023regtoken}, sharing the same embedding dimension, are concatenated with the patch tokens to encapsulate global semantic and redundant background information, respectively. We leverage these 2,080 (1,024-dimensional) output tokens as representation features for the corresponding image patches in the subsequent feature matching process, facilitating precise and accurate registration.

\subsection{Volumetric Feature Encoding}
\label{sec:vol_feature_extract}

To encode 3D medical images with the 2D DINOv2 encoder, we select one of the three orthogonal views (\emph{i.e.}, axial, coronal, and sagittal views) and encode all the slices in that view. Our experiments showed that the axial view works the best, which is thus our default choice. The shape of DINOv2 input patches are fixed at 14$\times$14, which means every 14$\times$14 patch is encoded into a feature vector. That will significantly limit the precision of image registration. We thus up-sample the input images by three times to obtain features with finer resolution. Each 2D slice from the image will be encoded into a 3D feature map with DINOv2, with the third dimension being the feature dimension. For 3D volumetric encoding, the output is 4D with an additional feature dimension. The 4D feature maps is obtained by stacking the 3D feature maps from all slices. After obtaining two 4D feature maps, $F_{\text{ref}}$ and $F_{\text{mov}}$, we perform principal component analysis (PCA) on all feature tokens in the two feature maps to reduce feature dimension. Let the desired feature length be $k$, the process is denoted as  
\begin{equation}
f,m = \text{PCA}(\left[F_{\text{ref}}, F_{\text{mov}}\right], k),
\end{equation}

where $k=24$ in this project. The $f$ and $m$ are the reference and moving feature maps used for registration. Due to the expensive time cost for encoding every slice in the image volume, we only encode every three slices and interpolate the feature maps in between. To further speed up the process, we swapped the full PCA with low rank PCA \cite{halko2011finding}, which significantly reduces run time with almost identical results. We have empirically validated our above design choices, including the choice of the encoding view, reduction of feature dimension using PCA, and slice interpolation interval. More details are explained in the ablation study section (Sec.~\ref{sec:ablation_study}).

\subsection{Discrete Optimization with Feature-wise LCC}
\label{sec:ncc_optimization}

Fast discrete registration has been previously investigated by Siebert et al \cite{siebert2021fast}. We adopted their method in splitting the optimization into two separate steps. In the first step the features are used to compute an SSD cost volume to give an initial registration field. Since optimizing the similarity metric subject to the smoothness constraint results in a non-convex optimization process, this field is then alternated on optimizing for similarity and smoothness\cite{steinbrucker2009large}. The resulting field is then used as the starting point for Adam-based optimization for more detailed local alignment.

While the reference and moving feature maps are roughly in the same feature space after PCA, there are still distribution differences between the features. This is especially true for noisy datasets, such as the CBCT dataset that our method is evaluated on. Therefore, instead of using the SSD (Sum of Squared Distance) as the loss function for the Adam-based optimization, we used feature-wise LCC (Local Cross Correlation) proposed in \cite{balakrishnan2019voxelmorph} to measure the similarity. The LCC is written as: 
\begin{equation}
LCC(f, m \circ \phi) = \sum_{p \in \Omega}  \frac{\left( \sum_{p_i} (f(p_i) - \hat{f}(p)) (m \circ \phi(p_i) - \hat{m \circ \phi}(p)) \right)^2}{\left( \sum_{p_i} (f(p_i) - \hat{f}(p))^2 \right) \left( \sum_{p_i} (m \circ \phi(p_i) - \hat{m \circ \phi}(p))^2 \right)},
\end{equation}
where $f(p_i)$ denotes each reference image feature, $\phi$ is the registration deformation field, $m \circ \phi(p_i)$ indicates each warped image feature, and $\hat{f}(p)$ $\hat{m \circ \phi}(p)$ are local mean image features. The LCC is less sensitive to the absolute difference between feature vectors, and better quantifies the trend and correlation across feature maps.

\section{Experiments and Results}

\subsection{OncoReg Challenge Dataset}

In this paper, we showcase the outcomes achieved by our method in the OncoReg challenge. The OncoReg challenge is categorized as a Type 3 challenge, signifying that the evaluation is conducted on a dataset that remains undisclosed to participants prior to the assessment. The participating teams submit their methods as Docker packages and the challenge organizers carry out the benchmarking process at their end without disclosing the test data. 
An auxiliary dataset, referred to as ThoraxCBCT, was made available to the participants, featuring a similar data structure. Both datasets focus on the intra-patient registration task, aligning pre-treatment fan-beam CT (FBCT) images with low-dose cone beam CT (CBCT) scans. The CBCT images are characterized by a reduced field of view and a significantly lower signal-to-noise ratio compared to FBCT images. The ThoraxCBCT dataset contains 20 pairs of images for training and 6 pairs for validation. Manual segmentations and landmarks are only available for the validation set for participants to evaluate and adjust their algorithms. 

In Section \ref{sec:challenge_results}, we discuss the benchmarking results from the concealed OncoReg challenge dataset, which are impartially compiled by the challenge organizers. Furthermore, in Section \ref{sec:ablation_study}, we delve into the findings of our experimental studies on the ThoraxCBCT dataset, systematically examining the impact of various components within our registration framework.

\subsection{Results}
\label{sec:challenge_results}

\begin{table}[t]
\centering
\caption{OncoReg challenge results. The score column is calculated by the challenge organizers to holistically represent the performance of each method.}
\label{tab:oncoreg_results}
\begin{tabular}{l|lrll|c|r}
\toprule

\textbf{Team} & TRE & TRE30 & DICE & sdLogJ & Score & Runtime \\
\midrule
\textbf{DINO-RegEn (ours)} & 3.5094 & 6.6556 & 0.6225  & 0.0394 & \textbf{0.742} & $>$ 300s \\
Voxelmorph++ \cite{heinrich2022voxelmorph++} & 3.7145 & 6.6213 & 0.6364 & 0.0683 & 0.717 & $<$ 60s \\
ConvexAdam \cite{siebert2021fast} & 3.4674 & 6.3387 & 0.6095  & 0.0588 & 0.686 & ($<$ 5s) \\
\textbf{DINO-Reg (ours)} & 3.8825 & 7.2551 & 0.5983  & 0.0310 & 0.656 & $>$ 300s \\
FourierNet & 4.7692 & 8.1065 & 0.6193  & 0.0945 & 0.544 & $>$ 300s \\
deedsBCV \cite{heinrich2013mrf} & 7.7359 & 10.6175 & 0.5793 & 0.1497 & 0.480 & $<$ 15s \\
NiftyReg \cite{rueckert1999nonrigid} & 7.4666 & 10.4846 & 0.3817  & 0.0531 & 0.378 & $<$ 60s \\
zerofield (No Registration) & 6.0679 & 10.1208 & 0.4423 & - & - & - \\
\bottomrule
\end{tabular}
\end{table}

The OncoReg challenge uses a combination of multiple metrics, including TRE (Target Registration Error), TRE30 (TRE of landmarks with 30 percentile largest initial error), DICE, and sdLogJ (Standard deviation of log Jacobian determinant of the registration field), to rank the submissions. The coefficient of each evaluation metric towards the final rank was determined by the challenge organizers. Table~\ref{tab:oncoreg_results} shows the final challenge leaderboard in descending order.
Both DINO-RegEn (\#1 place) and DINO-Reg (\#4 place) are the submissions from our team. DINO-Reg in the fourth row of Table~\ref{tab:oncoreg_results} represents the native result of the algorithm as described in Section~\ref{sec:method}. DINO-RegEn is the ensemble with ConvexAdam as detailed below. 

\begin{figure}[t]
\includegraphics[width=\textwidth]{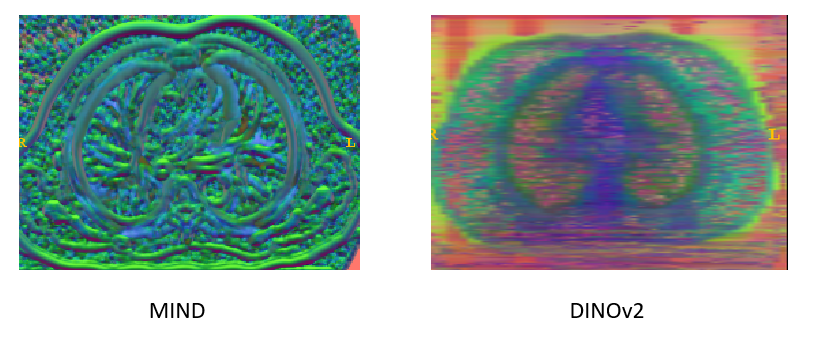}
\caption{Comparison between MIND features and DINOv2 features. First 3 principal components of each feature are visualized as RGB image. Visualization created with \cite{py06nimg}.} \label{fig2_feature}
\end{figure}

In our experiments with the ThoraxCBCT datset, we observe that the ConvexAdam method \cite{siebert2021fast}, which shares the same optimization scheme with DINO-Reg but with different loss function and feature encoder (MIND\_SSC), to be slightly better in DICE and lower in TRE. Comparing the DINOv2 features with MIND features in Figure~\ref{fig2_feature}, we found that although DINOv2 features contain much richer semantics, it is less enunciated on the corner points and contours of organs. When registering fine-grained details, such ambiguity on contours leads to misalignment. Therefore, we also present DINO-Reg\_ensemble, which takes the result of ConvexAdam \cite{siebert2021fast} and DINO-Reg and computes the mean registration field from the two algorithms. Our premise is that the detailed local gradient and intensity-based features of MIND complement the rich global semantics of DINOv2 features. The result in Table~\ref{tab:oncoreg_results} supports our hypothesis. In later experiments shown in Table~\ref{tab:ensemble}, we found that performing registration with DINO-Reg and MIND+Adam sequentially, which is equivalent to using DINO-Reg as the global registration step and MIND+Adam as the local finetuning step, yields the highest DICE on the ThoraxCBCT dataset. MIND+Adam is different from ConvexAdam in omitting the global convex optimization step. This result further proves our hypothesis that using DINO-Reg and MIND features as sequential global and finetuning registration steps is a logical approach. Note that reversing the order results in worse results than using DINO-Reg alone.

Additionally, after the challenge submission, we optimized the loss function implementation, adopted low-rank PCA, and used interpolated slice encoding to accelerate our algorithm. The runtime of our methods has been reduced from more than 300 seconds (s) to $\sim$60s per case. 

\begin{table}[t]
\centering
\caption{Comparison between the results using MIND and DINOv2 features under various scenarios of ensembles. MIND+Adam stands for ConvexAdam without the global convex optimization step. Experiments were performed on the ThoraxCBCT dataset. }
\label{tab:ensemble}
\begin{tabular}{l|ccc}
\toprule
Method & TRE & LogJacDetStd & Dice\_mean$\pm$std \\ \midrule
DINO-Reg & $3.86 \pm 0.93$ & $0.038 \pm 0.005$ & $0.724 \pm 0.15$ \\
ConvexAdam \cite{siebert2021fast} & $4.18 \pm 0.36$ & $0.076 \pm 0.018$ & $0.762 \pm 0.13$ \\
DINO-RegEn (DINO-Reg+ConvexAdam) & $3.72 \pm 0.68$ & $0.047 \pm 0.011$ & $0.753 \pm 0.14$ \\
ConvexAdam then DINO-Reg & $3.92 \pm 0.93$ & $0.038 \pm 0.004$ & $0.725 \pm 0.15$ \\
DINO-Reg then MIND+Adam & $3.92 \pm 0.76$ & $0.066 \pm 0.013$ & $\mathbf{0.771 \pm 0.13}$ \\ \bottomrule
\end{tabular}
\end{table}

\subsection{Ablation Studies}
\label{sec:ablation_study}

\begin{figure}[t]
\includegraphics[width=\textwidth]{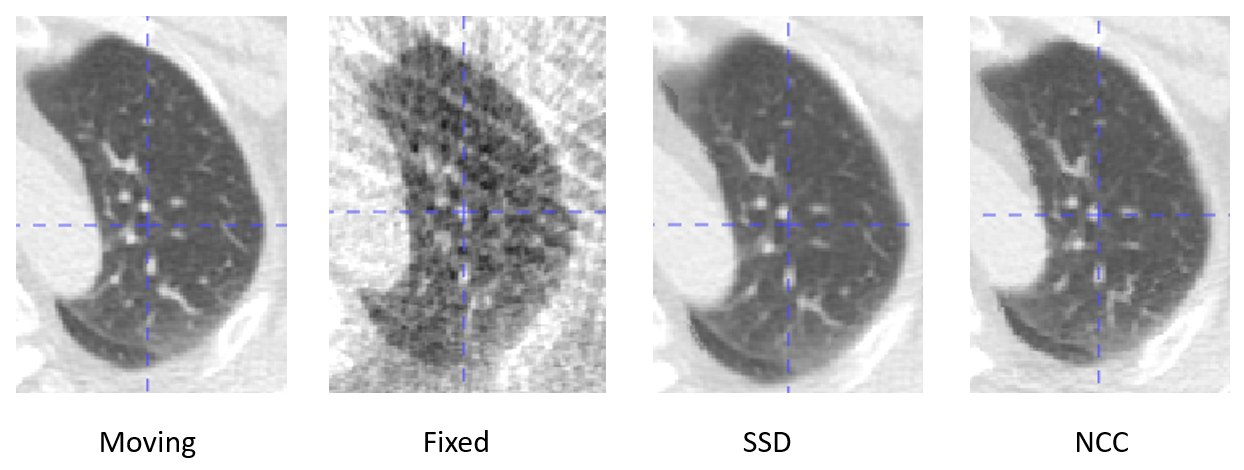}
\caption{Qualitative comparison between registration results from DINO-Reg with different loss functions during convex optimization. The cross hair is located at the same physical location across all images.} \label{fig3_quailtative}
\end{figure}

In this section, we discuss the rationale for choosing our loss function (NCC vs. SSD), hyperparameter setting (number of slices to skip during encoding), and the encoding views. In Table~\ref{tab:ablation_results}, the first two rows offer a directly quantitative comparison between the performance of using SSD and NCC as loss functions, respectively, for the convex optimization. Using NCC yields significantly better result in both TRE and Dice. Figure~\ref{fig3_quailtative} also shows the qualitative registration result using the two different loss functions. In Figure~\ref{fig3_quailtative}, the crosshair marks the same physical location across all images. The structure marked by the cross hair in the fixed image is far from the corsshair in the moving image. After NCC-guided registration, the structure overlaps nicely with the cross hair.

In Table~\ref{tab:ablation_results}, rows 2 to 6 show the result of encoding every 1, 2, 3, or 5 slices and interpolating the features for the skipped slices in between. We can observe that the the performance is stable until the gap increases beyond 3. As a result, we use Gap=3 in our following experiments. Rows 4 and 5 also compare the difference between using full PCA and low rank PCA. Though the result is almost identical, the time consumption on this step is up to 50$\times$ less when using low rank PCA. 

\begin{table}[t]
\centering
\caption{Performance of DINO-Reg under various configurations.}
\label{tab:ablation_results}
\begin{tabular}{l|ccc|ccc}
\toprule
Row \# & Gap & PCA & Optimization & TRE & LogJacDetStd & Dice\_mean \\ 
\midrule
1 & 1 & normal & SSD; 800 epoch & $4.54 \pm 0.86$ & $0.018 \pm 0.002$ & $0.732 \pm 0.15$ \\
2 & 1 & normal & NCC; 50 epoch & $3.87 \pm 0.87$ & $0.050 \pm 0.002$ & $0.732 \pm 0.15$ \\
3 & 2 & normal & NCC; 50 epoch & $3.88 \pm 0.86$ & $0.045 \pm 0.003$ & $0.728 \pm 0.15$ \\
4 & 3 & normal & NCC; 50 epoch & $3.86 \pm 0.83$ & $0.044 \pm 0.003$ & $0.726 \pm 0.15$ \\
5 & 3 & low rank & NCC; 50 epoch & $3.86 \pm 0.82$ & $0.044 \pm 0.003$ & $0.727 \pm 0.15$ \\
6 & 5 & normal & NCC; 50 epoch & $3.96 \pm 0.80$ & $0.044 \pm 0.003$ & $0.709 \pm 0.16$ \\
\bottomrule
\end{tabular}
\end{table}

\begin{table}[t]
\centering
\caption{Comparison of DINOv2 feature performance with different orientations. A experiments in this table use Gap = 3, low rank PCA, NCC loss function, 50 epochs. The last column shows the percentge of varicance captured by the first 3 principal components.}
\label{tab:encoding_views}
\begin{tabular}{l|cccc}
\toprule
Encoding View  & TRE & LogJacDetStd & Dice\_mean & \% Variance PCA\_3\\ \midrule
Coronal  & $3.86 \pm 0.82$ & $0.044 \pm 0.003$ & $0.727 \pm 0.15$ & 40.5\%\\
Sagittal & $4.03 \pm 0.66$ & $0.038 \pm 0.003$ & $0.714 \pm 0.15$ & 38.8\% \\
Axial  & $3.84 \pm 0.88$ & $0.043 \pm 0.004$ & $\mathbf{0.733 \pm 0.14}$ & 43.6\% \\
\bottomrule
\end{tabular}
\end{table}

Table~\ref{tab:encoding_views} illustrates the performance of the DINO-Reg with different encoding views. The results indicate that the axial view yields the highest Dice mean score of 0.733 and a competitive TRE. We also observe that the first 6 principal components in axial view capture the most amount of variance and coronal view comes in second, while the sagittal view captures the least. This hierarchy in performance and variance capture is consistent with the findings presented in Table \ref{tab:encoding_views}.

\section{Conclusion}
In this paper we present a training-free algorithm, DINO-Reg. When we combine the strengths of both general purpose deep learning features and handcrafted features, the registration result is robust even on unseen datasets, as proven in the OncoReg challenge. The success of DINO-Reg highlights the potential for general vision foundation models to contribute significantly to medical image analysis. In the future, we expect to see similar innovations bridging the gap between medical image and general vision.

%
%
%
\bibliographystyle{splncs04}
\bibliography{reference}

\begin{thebibliography}{10}
\providecommand{\url}[1]{\texttt{#1}}
\providecommand{\urlprefix}{URL }
\providecommand{\doi}[1]{https://doi.org/#1}

\bibitem{baharoon2023dinoradiology}
Baharoon, M., Qureshi, W., Ouyang, J., Xu, Y., Aljouie, A., Peng, W.: Towards general purpose vision foundation models for medical image analysis: An experimental study of {DINOv2} on radiology benchmarks. arXiv preprint arXiv:2312.02366  (2023)

\bibitem{balakrishnan2019voxelmorph}
Balakrishnan, G., Zhao, A., Sabuncu, M.R., Guttag, J., Dalca, A.V.: Voxelmorph: a learning framework for deformable medical image registration. IEEE transactions on medical imaging  \textbf{38}(8),  1788--1800 (2019)

\bibitem{caron2021dinov1}
Caron, M., Touvron, H., Misra, I., J{\'e}gou, H., Mairal, J., Bojanowski, P., Joulin, A.: Emerging properties in self-supervised vision transformers. In: Proceedings of the IEEE/CVF international conference on computer vision. pp. 9650--9660 (2021)

\bibitem{darcet2023regtoken}
Darcet, T., Oquab, M., Mairal, J., Bojanowski, P.: Vision transformers need registers. arXiv preprint arXiv:2309.16588  (2023)

\bibitem{deng2009imagenet}
Deng, J., Dong, W., Socher, R., Li, L.J., Li, K., Fei-Fei, L.: Imagenet: A large-scale hierarchical image database. In: 2009 IEEE conference on computer vision and pattern recognition. pp. 248--255. Ieee (2009)

\bibitem{dosovitskiy2020vit}
Dosovitskiy, A., Beyer, L., Kolesnikov, A., Weissenborn, D., Zhai, X., Unterthiner, T., Dehghani, M., Minderer, M., Heigold, G., Gelly, S., et~al.: An image is worth 16x16 words: Transformers for image recognition at scale. arXiv preprint arXiv:2010.11929  (2020)

\bibitem{halko2011finding}
Halko, N., Martinsson, P.G., Tropp, J.A.: Finding structure with randomness: Probabilistic algorithms for constructing approximate matrix decompositions. SIAM review  \textbf{53}(2),  217--288 (2011)

\bibitem{haskins2019learning}
Haskins, G., Kruecker, J., Kruger, U., Xu, S., Pinto, P.A., Wood, B.J., Yan, P.: Learning deep similarity metric for {3D} {MR--TRUS} image registration. International journal of computer assisted radiology and surgery  \textbf{14},  417--425 (2019)

\bibitem{haskins2020deep}
Haskins, G., Kruger, U., Yan, P.: Deep learning in medical image registration: a survey. Machine Vision and Applications  \textbf{31},  1--18 (2020)

\bibitem{heinrich2022voxelmorph++}
Heinrich, M.P., Hansen, L.: Voxelmorph++ going beyond the cranial vault with keypoint supervision and multi-channel instance optimisation. In: International Workshop on Biomedical Image Registration. pp. 85--95. Springer (2022)

\bibitem{heinrich2012mind}
Heinrich, M.P., Jenkinson, M., Bhushan, M., Matin, T., Gleeson, F.V., Brady, M., Schnabel, J.A.: Mind: Modality independent neighbourhood descriptor for multi-modal deformable registration. Medical image analysis  \textbf{16}(7),  1423--1435 (2012)

\bibitem{heinrich2013mrf}
Heinrich, M.P., Jenkinson, M., Brady, M., Schnabel, J.A.: Mrf-based deformable registration and ventilation estimation of lung ct. IEEE transactions on medical imaging  \textbf{32}(7),  1239--1248 (2013)

\bibitem{hering2022learn2reg}
Hering, A., Hansen, L., Mok, T.C., Chung, A.C., Siebert, H., H{\"a}ger, S., Lange, A., Kuckertz, S., Heldmann, S., Shao, W., et~al.: Learn2reg: comprehensive multi-task medical image registration challenge, dataset and evaluation in the era of deep learning. IEEE Transactions on Medical Imaging  \textbf{42}(3),  697--712 (2022)

\bibitem{kingma2014adam}
Kingma, D.P., Ba, J.: Adam: A method for stochastic optimization. arXiv preprint arXiv:1412.6980  (2014)

\bibitem{li2023samconvex}
Li, Z., Tian, L., Mok, T.C., Bai, X., Wang, P., Ge, J., Zhou, J., Lu, L., Ye, X., Yan, K., et~al.: Samconvex: Fast discrete optimization for ct registration using self-supervised anatomical embedding and correlation pyramid. In: International Conference on Medical Image Computing and Computer-Assisted Intervention. pp. 559--569. Springer (2023)

\bibitem{mok2020large}
Mok, T.C., Chung, A.C.: Large deformation diffeomorphic image registration with laplacian pyramid networks. In: Medical Image Computing and Computer Assisted Intervention--MICCAI 2020: 23rd International Conference, Lima, Peru, October 4--8, 2020, Proceedings, Part III 23. pp. 211--221. Springer (2020)

\bibitem{oquab2023dinov2}
Oquab, M., Darcet, T., Moutakanni, T., Vo, H., Szafraniec, M., Khalidov, V., Fernandez, P., Haziza, D., Massa, F., El-Nouby, A., et~al.: {DINOv2}: Learning robust visual features without supervision. arXiv preprint arXiv:2304.07193  (2023)

\bibitem{rueckert1999nonrigid}
Rueckert, D., Sonoda, L.I., Hayes, C., Hill, D.L., Leach, M.O., Hawkes, D.J.: Nonrigid registration using free-form deformations: application to breast mr images. IEEE transactions on medical imaging  \textbf{18}(8),  712--721 (1999)

\bibitem{siebert2021fast}
Siebert, H., Hansen, L., Heinrich, M.P.: Fast {3D} registration with accurate optimisation and little learning for {Learn2Reg} 2021. In: International Conference on Medical Image Computing and Computer-Assisted Intervention. pp. 174--179. Springer (2021)

\bibitem{song2022cross}
Song, X., Chao, H., Xu, X., Guo, H., Xu, S., Turkbey, B., Wood, B.J., Sanford, T., Wang, G., Yan, P.: Cross-modal attention for multi-modal image registration. Medical Image Analysis  \textbf{82},  102612 (2022)

\bibitem{song2021cross}
Song, X., Guo, H., Xu, X., Chao, H., Xu, S., Turkbey, B., Wood, B.J., Wang, G., Yan, P.: Cross-modal attention for mri and ultrasound volume registration. In: Medical Image Computing and Computer Assisted Intervention--MICCAI 2021: 24th International Conference, Strasbourg, France, September 27--October 1, 2021, Proceedings, Part IV 24. pp. 66--75. Springer (2021)

\bibitem{steinbrucker2009large}
Steinbr{\"u}cker, F., Pock, T., Cremers, D.: Large displacement optical flow computation withoutwarping. In: 2009 IEEE 12th International Conference on Computer Vision. pp. 1609--1614. IEEE (2009)

\bibitem{ye2022desd}
Ye, Y., Zhang, J., Chen, Z., Xia, Y.: {DeSD}: Self-supervised learning with deep self-distillation for {3D} medical image segmentation. In: International Conference on Medical Image Computing and Computer-Assisted Intervention. pp. 545--555. Springer (2022)

\bibitem{py06nimg}
Yushkevich, P.A., Piven, J., Cody~Hazlett, H., Gimpel~Smith, R., Ho, S., Gee, J.C., Gerig, G.: User-guided {3D} active contour segmentation of anatomical structures: Significantly improved efficiency and reliability. Neuroimage  \textbf{31}(3),  1116--1128 (2006)

\end{thebibliography}

\end{document}